\documentclass[runningheads]{llncs}

\usepackage{graphicx}
\usepackage[table]{xcolor}
\newcommand{\etal}{\textit{et al.}}
\usepackage{hyperref}

\begin{document}


\title{A Principled Approach to \\ Failure Analysis and Model Repairment: Demonstration in Medical Imaging}
\titlerunning{A Principled Approach to Failure Analysis and Model Repairment}
%
\author{Thomas Henn\inst{1} \and
Yasukazu Sakamoto\inst{2} \and
Clément Jacquet\inst{1} \and
Shunsuke~Yoshizawa\inst{2}\and
Masamichi Andou\inst{2} \and
Stephen Tchen\inst{1} \and
Ryosuke~Saga\inst{1} \and
Hiroyuki~Ishihara\inst{2} \and 
Katsuhiko Shimizu\inst{2} \and 
Yingzhen Li\inst{3} \and
Ryutaro Tanno\inst{4}}

\institute{Rokken Inc., \email{thomas.henn@lab6.co.jp}
\and
Terumo Corporation, Corporate R\&D Center,
\email{yasukazu\_sakamoto@terumo.co.jp}\\
 \and
Department of Computing, Imperial College London\\
 \and
Department of Computing, University College London\\}

\authorrunning{T. Henn et al.}
%
%
\maketitle              

\begin{abstract}
Machine learning models commonly exhibit unexpected failures post-deployment due to either data shifts or uncommon situations in the training environment.
Domain experts typically go through the tedious process of inspecting the failure cases manually, identifying failure modes and then attempting to fix the model.
In this work, we aim to standardise and bring principles to this process through answering two critical questions: (i) how do we know that we have identified meaningful and distinct failure types?; (ii) how can we validate that a model has, indeed, been repaired?
We suggest that the quality of the identified failure types can be validated through measuring the intra- and inter-type generalisation after fine-tuning and introduce metrics to compare different subtyping methods. Furthermore, we argue that a model can be considered repaired if it achieves high accuracy on the failure types while retaining performance on the previously correct data.
We combine these two ideas into a principled framework for evaluating the quality of both the identified failure subtypes and model repairment. We evaluate its utility on a classification and an object detection tasks.
Our code is available at \url{https://github.com/Rokken-lab6/Failure-Analysis-and-Model-Repairment}

\end{abstract}


\section{Introduction}

It is common for new failures to be discovered once a model has been deployed in the ``wild''. While recent lines of research in medical imaging have shown promising results in designing robust machine learning (ML) models \cite{panfilov2019improving,bdair2020roam,billot2020learning,liu2020ms,dou2019domain}, it may not be realistic to achieve perfect generalisation to every relevant environment.

Consequently, recently published guidelines for the reliable application of ML systems in healthcare~\cite{collins2019reporting,liu2020reporting} and recent work from Luke \etal \cite{oakden2020hidden} stress the importance of analyzing and reporting clinically relevant failure cases.
However, there is a lack of standardised protocols to identify, validate and analyze those failure types. Typically, domain experts manually inspect the failure cases and make sense of them by identifying a set of failure modes. But this approach can be both expensive and biased by the human expertise. For example, a critical yet rare subgroup could be missed with such an approach and go unreported \cite{oakden2020hidden}. A notable recent work \cite{singlaCVPR2021} recognises this issue and makes a first step towards data-driven approaches to failure subtyping through clustering of the learned features based on whether its presence or absence is predictive of poor performance.
However to date, little attention has been gathered around the evaluation metrics of the identified failure types, hampering the development of new methods in this direction. 
Furthermore, even if a set of meaningful failure types could be identified, methods for \textit{fixing} them and evaluating its success remain undeveloped.

In an attempt to bring principles to the process of failure analysis and model repairment, we introduce a framework for not only deriving subtypes of failure cases and measuring their quality, but also repairing the models and verifying their generalisation. 
We put forward a set of desirable properties that a meaningful set of failure subtypes should meet and design surrogate metrics.
Moreover, we propose a data-driven method for identifying failure types based on clustering in feature or gradient spaces.
This method was able not only to identify failure types with the highest quality according to our metrics but also to identify clinically important failures like undetected catheters close to the ultrasound probe in intracardiac echocardiography.
Finally, we argue that model repairment should not only aim to fix each failure type in a generalizable manner but also to ensure the performance on previously successful cases is retained.

\begin{figure}[ht!]
    \centering
  \label{fig:pipeline}
    
    \includegraphics[width=\linewidth]{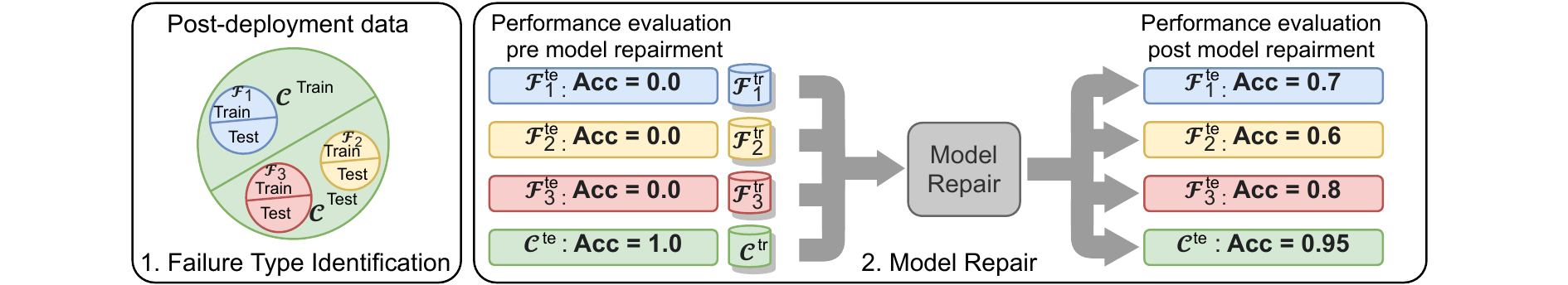}

    \caption{An overview of the proposed failure analysis and model repairment framework which proceeds in two phases.
Firstly, based on the evaluation of the model $\mathcal{M}$, the set of ``failure cases'' $\mathcal{F}$ (e.g., misclassified cases, examples with high errors, etc) are clustered to identify failures types $\{\mathcal{F}_i\}_{i}$ such that $\mathcal{F} = \cup_{i}\mathcal{F}_i$.
Secondly, the failure types $\{\mathcal{F}_i\}_{i}$ are ``fixed'' by a \textit{repairment} algorithm (e.g., fine-tuning, continual learning methods, etc) based on a ``training split'' of the failure sets $\{\mathcal{F}_i^{tr}\}$, and the correct set $\mathcal{C}^{tr}$. 
Finally, the success of the repairment is evaluated by measuring generalisation on the test sets of target failure types $\{\mathcal{F}_i^{te}\}$ and the correct cases $\mathcal{C}^{te}$.}
    
\end{figure}


\section{Methods}

We elaborate below the two phases (see Fig.~\ref{fig:pipeline}) of our approach to model repair.

\subsection{Phase I: identification of failure types}

It is not evident how to optimally separate failure cases into a set of distinct failure types. 
Domain experts could split the failure cases according to the visual appearance or consider the importance of different failures from a clinical perspective, for example according to stages or clinical signs of a disease. 
However, we suggest clinical relevance doesn't necessarily reflect the way the model distinctly fails on each failure type.
Failure types should be specific not only to the model but the repairment methods that can be used to fix them.
Moreover, objective metrics to quantify the quality of failure types for the purpose of model repairment are still lacking.

\subsubsection{Two desiderata of failure types: }We postulate that a set of failure types should satisfy two desirable properties: \textit{Independence} and \textit{Learnability}.
In addition, we propose two novel surrogate metrics to assess those properties in practice. They are measured by fine-tuning the model $\mathcal{M}$ on a subset $\mathcal{F}_i^{tr}$ of the given failure type $\mathcal{F}_i$ under a \textit{Compatibility constraint}, and then calculating the performance on each $\mathcal{F}_{j}^{te}$.
We assume a sufficiently large set of failures with no label noise or corrupted images.

\textit{(i) Independence:} The subtypes should be as independent as possible from each other in the way the model fails. In other words, they can be fixed by comparatively distinct alterations to the decision boundary.
If two types are independent, a model fine-tuned on one type should not be useful to the other.
In practice, we suggest a continuous measure of \textit{Independence} by calculating the average difference in the test performance between each failure type and the rest:
$I(\mathcal{F}_{i}) := \mathrm{mean}_{j\neq i}\left( m(\mathcal{M}_i, \mathcal{F}_{i}^{te})  - m(\mathcal{M}_i, \mathcal{F}_{j}^{te}) \right )$
where $m(\cdot)$ is a performance metric (e.g., accuracy).

\textit{(ii) Learnability:} Each subtype should be homogeneous and consists of examples for which the model has failed in a similar way. In other words, a such failure type contains failure cases that can be fixed via a similar modification to the model's decision boundary. 
If a subtype is heterogeneous, fixing it would require more modifications to the model and would be more challenging.
We thus posit that a more homogeneous failure type would be easier to learn, and measure \textit{Learnability} as the generalisation of the fine-tuned model $\mathcal{M}_i$ on the chosen failure type $
L(\mathcal{F}_{i}) :=  m(\mathcal{M}_i, \mathcal{F}_{i}^{te}) 
$.

\textit{Compatibility constraint:} We argue in addition that the above surrogate metrics should be measured with the constraint of maintaining performance on correct data $\mathcal{C}$.
This is necessary to avoid learning pathological discriminative rules just to solve a specific failure type. For example, given a failure type only containing images of a single class, the model could simply learn to ignore the input images and predict the same class everywhere, in which case the failure is fixed in a meaningless way. \textit{Compatibility} ensures ``locality'' of the failure types by ensuring the required changes in the discriminative rules do not considerably influence the previous cases.
\textit{Compatibility} is achieved by fine-tuning on both a failure type $\mathcal{F}_i^{tr}$ and previously correct cases $\mathcal{C}^{tr}$ in equal proportions and early stopping when $\mathcal{C}^{tr}$ validation performance drops below 0.9.

\subsubsection{Automatic identification of failure types:}
Manual analysis can often be time-consuming but also suboptimal, potentially overlooking meaningful failure modes.
Therefore, we wish to automatically uncover a set of failure types with good \textit{Independence} and \textit{Learnability} scores.
We thus also explore two methods as specific instantiations of our framework;
particularly, we experiment with clustering the failure cases $\mathcal{F}$ in the feature space (\textit{Feature clustering}) and gradient space of a differentiable model $\mathcal{M}$ (\textit{Gradient clustering}).
The gradients of the loss with respect to the parameters are used (for object detection: the loss specific to the object).
We expect that similar data will be close in the feature space and the failures whose correction requires similar changes to the model parameters will be close in the gradient space.
Furthermore, features are averaged over spatial dimensions and both features and gradients are reduced by Gaussian random projection followed by UMAP \cite{mcinnes2018umap-software} to 10 dimensions.
Finally, the data is clustered through k-means with the highest Silhouette score between 3 and 10 clusters.

\subsection{Phase II: repairment of failure types}

Once a set of target failure types has been identified, the model needs to be repaired. We argue that a successful repairment leads to a model that generalizes on unseen cases of the target failure types while maintaining performance on the cases $\mathcal{C}$ where the model previously performed well.
First, a target set of failure types $\mathcal{F}^{T} \subset \{\mathcal{F}_i\}_{i}$ is selected by the end-users, and then each type is split into two sets, where one is used to first repair the model (e.g. $\mathcal{F}_i^{tr}$) and then the other is used to evaluate generalization (e.g. $\mathcal{F}_i^{te}$). For repairment procedures, we experiment with fine-tuning on the failure types (with or without the correct cases) and \textit{elastic weight consolidation} (EWC), a popular continual learning approach \cite{kirkpatrick2017overcoming}. We note, however, that more recent model adaptation approaches are also applicable in our framework such as more recent variants of continual learning \cite{karani2018lifelong,hofmanninger2020dynamic}, meta-learning \cite{karani2021test} and domain adaptation \cite{kamnitsas2017unsupervised}.


\section{Datasets and Implementation Details}

We evaluate the efficacy of the proposed model repairment framework on two
medical imaging datasets. The details of the respective datasets along with the specification of models/optimisation are provided below.

\textbf{Binary PathMNIST (BP-MNIST)} is a publicly available classification dataset consisting of colorectal cancer histology slides patches \cite{medmnist,kather2019predicting} derived originally from the NCT-CRC-HE-100K~\cite{kather2019predicting} dataset and resized to 3x28x28 as a part of the MedMNIST benchmark~\cite{medmnist}. We simplify the original 9-way classification task into a binary task of discriminating benign and malignant classes (\textit{cancer-associated stroma} and \textit{colorectal adenocarcinoma epithelium}) and use the original classes of granular tissue types as metadata to interpret the discovered failure types.
Moreover, 40\% of the dataset was put into the test set to increase the sample size for the evaluation of both subtyping and model repairment. Finally, the model was trained with Adam with a learning rate of $10^{-4}$ in combination with early stopping on the validation accuracy. The architecture is a version of VGG \cite{simonyan2014veryVGG} with 6 convolutional layers starting at 16 channels and the fully connected layer replaced by a 1x1 convolution to two output channels and a spatial average.

\textbf{ICE Catheter Detection (ICE-CD)} is a private real-world object detection dataset comprised of ultrasound images of intra-cardiac catheters made by a Intracardiac Echocardiography (ICE) device on pigs.
Furthermore, for the purpose of evaluating the performance of catheter detection models, each catheter image was classified into different types representing known difficult situations based on catheter appearance or position. In addition, information about the rough anatomical locations of the probe is available as metadata. The architecture is composed of 5 residual blocks of two convolutional layers (starting at 8 channels and doubling up to 128 channels) followed by two 1x1 convolutions branches: a classification and a center position regression branch. This dataset has been acquired in accordance with animal experiment regulations.


\section{Experiments and Results}

\subsection{Comparison of failure subtyping methods}

\begin{figure}[ht!]
  \centering

    \includegraphics[width=\linewidth]{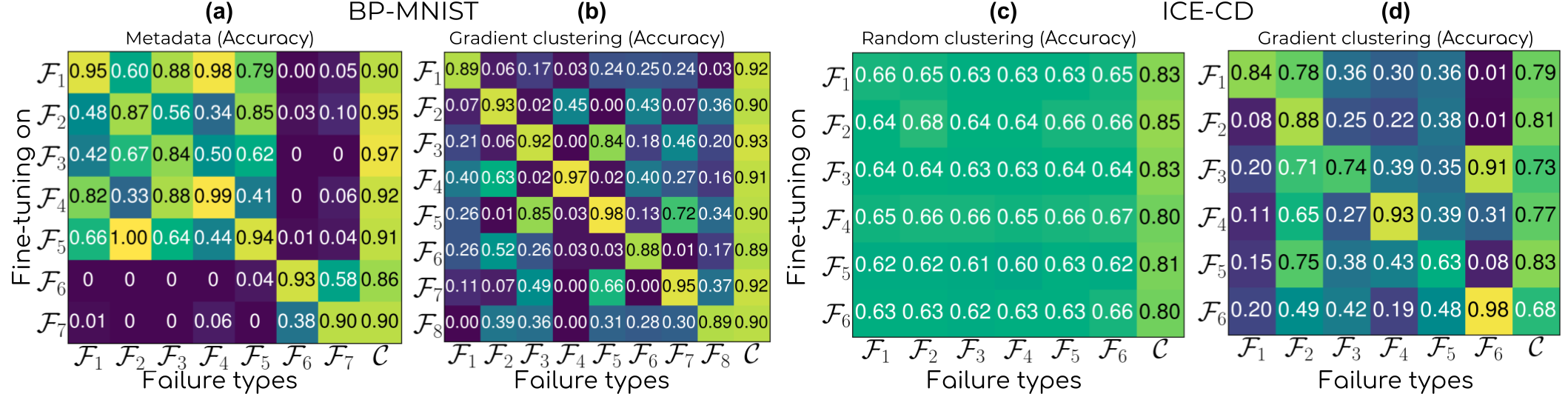}

    \caption{Accuracy of fine-tuning on each failure type on BP-MNIST according to tissue types (a) and gradient clustering (b) and on ICE-CD according to random clustering (c) and gradient clustering (d). The lack of independence is apparent for tissue types (a) and random clustering (c) while a diagonal pattern is noticeable in (b) and (d).}
    
    \label{fig:finetuning_matrices}
    
\end{figure}

\textbf{Baselines and Experiments:} we aim to quantify the quality of the proposed automatic subtyping methods (see Table~\ref{tab:evaluation_types} and Fig~\ref{fig:finetuning_matrices}) which we compare against several baselines:
\textit{Random} (random clusters), \textit{False positives and negatives (FP/FN)} (two clusters) and \textit{Metadata} (BP-MNIST: the original 9 \textit{tissue types}; ICE-CD: \textit{image types} as identified by the clinicians and in addition \textit{anatomical locations} of the images). For each failure type $\mathcal{F}_i$, the model was fine-tuned on both $\mathcal{F}_i^{tr}$ and the correct cases $\mathcal{C}^{tr}$ (to satisfy \textit{Compatibility}). Early stopping is performed with the best validation score (accuracy on BP-MNIST and $F_1$ on ICE-CD) on $\mathcal{F}_i^{tr}$ while maintaining $0.9$ validation accuracy on $\mathcal{C}^{tr}$. Table~\ref{tab:evaluation_types} displays the average metrics for the respective methods while Fig~\ref{fig:finetuning_matrices} shows granular results i.e., matrix $m(\mathcal{M}_i, \mathcal{F}_j^{te})$ which denotes the test accuracy on $\mathcal{F}_j^{te}$ of the model fine-tuned on $\mathcal{F}_i^{tr}$.

\textbf{Analysis:} first of all, \textit{Gradient clustering} reached better scores than any other method with the exception of \textit{Independence} for \textit{FP/FN} clustering on BP-MNIST. However, it had a 18\% higher Learnability score and is more informative with more identified failure types.
Remarkably, \textit{Gradient clustering} was better than using \textit{Metadata}, including the ICE-CD metadata made through very time consuming visual inspection. 
In the case of BP-MNIST, the lack of independence of \textit{Metadata} subtyping was clearly visible in Fig.~\ref{fig:finetuning_matrices}(a).
This implies that \textit{Gradient clustering} might be able to identify independent failure types which are not obvious through human eyes but are relevant to the model.
On the other hand, \textit{Feature clustering} seemed to achieve lower scores than metadata.
Furthermore, \textit{Gradient clustering} might achieve higher \textit{Learnability} and \textit{Independence} due to being more aligned with the repairment method.
Finally, \textit{Random} resulted in by far the lowest independence scores as is apparent in Fig.~\ref{fig:finetuning_matrices}(c) where all types had the same score. This shows that the \textit{Independence} metric is effective in detecting when failure types are mixed together.
Moreover, \textit{Random} and \textit{FP/FN} clustering showed lower \textit{Learnability} which may be explained by the diversity of tasks to be learned within each cluster. However, \textit{FP/FN} had the highest \textit{Independence} due to matching the two classes.

\begin{table}[ht!]
    \caption{Comparison of methods for failure types identification on both BP-MNIST and ICE-CD. The best and the second best results are shown in red and blue.}
    \label{tab:evaluation_types}
    \centering
    \begin{tabular}{|c|c|c|c|c|}
    \hline
    
    ~ &  \multicolumn{2}{c|}{BP-MNIST (ACC)} & \multicolumn{2}{c|}{Catheter Detection (ACC)}\\ 
    
    Method &  Learnability & Independence&  Learnability  & Independence \\ \hline
    
    Random & 0.42$\pm$0.01 & 0.00$\pm$0.02 &                  0.65$\pm$0.02 & 0.01$\pm$0.01 \\ \hline
    
    False positives and negatives &  0.74$\pm$0.22 & \cellcolor{red!15}\textbf{0.74$\pm$0.22}                 & 0.69$\pm$0.14 & 0.14$\pm$0.37 \\ \hline
    
    BP-MNIST: \textit{tissue type} &   \cellcolor{red!15} 0.92$\pm$0.05 & 0.58$\pm$0.16                & - & - \\ \hline
    
    ICE-CD: \textit{image type} &   - & -                & \cellcolor{blue!15}0.77$\pm$0.17 & \cellcolor{blue!15} 0.46$\pm$0.16 \\ \hline

    ICE-CD: \textit{anatomical location} &  - & -                                         & 0.75$\pm$0.22 & 0.37$\pm$0.20 \\ \hline
    
    Feature clustering &  \cellcolor{blue!15}0.79$\pm$0.25 & 0.52$\pm$0.23                            & \cellcolor{red!15}\textbf{0.83$\pm$0.06} & 0.36$\pm$0.07 \\ \hline
    
    Gradient clustering & \cellcolor{red!15} \textbf{ 0.92$\pm$0.04}  &\cellcolor{blue!15} 0.69$\pm$0.04      & \cellcolor{red!15}\textbf{0.83$\pm$0.12} & \cellcolor{red!15} \textbf{0.48$\pm$0.18} \\ \hline
    
    \end{tabular}
\end{table}

\subsection{Analysis of automatically discovered failure subtypes}
We aim to inspect the automatically discovered subtypes on BP-MNIST and ICE-CD by \textit{Gradient clustering} which achieved the best scores.

\begin{figure}[ht!]
    \centering
    
    \includegraphics[width=\linewidth]{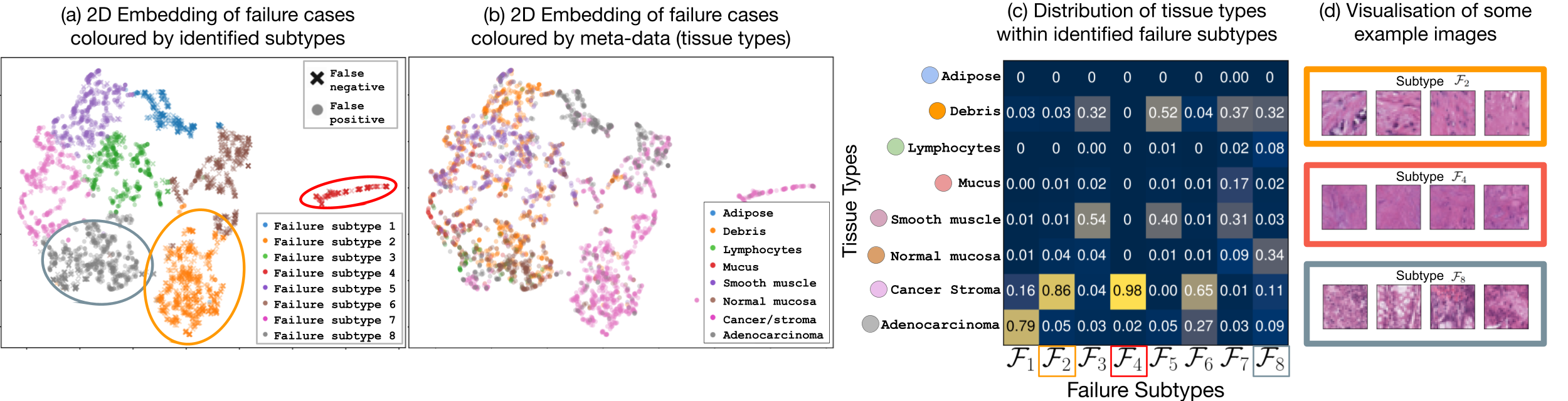}

    \caption{Inspection of failure types obtained through gradient space clustering for BP-MNIST. (a) UMAP embedding coloured by \textit{failure types}. (b) UMAP embedding coloured by \textit{tissue types}. (c) Distribution of tissue types within each failure type. (d) Example images for remarkable failure types.  Failures $\mathcal{F}_2$ and $\mathcal{F}_4$ which contain the same tissue but different visual appearances are identified as two different types.}

  \label{fig:cluster_analysis_pathmnist}
\end{figure}

\textbf{Binary PathMNIST (BP-MNIST)}: first, we observe that False positives and false negatives were mostly separated into two sets of clusters (i.e., each cluster contains mostly either circles or crosses as shown in Fig.~\ref{fig:cluster_analysis_pathmnist}(a)).
Moreover, the two malignant tissue types were recovered separately: Cluster 2 and 4 for \textit{cancer-associated stroma} and cluster 1 for \textit{colorectal adenocarcinoma epithelium}.
Secondly, even within one tissue type, \textit{Gradient clustering} was able to discover independent failure types.
Cluster 2 and 4 both focused on \textit{cancer-associated stroma} but were relatively independent and differed when evaluating on $\mathcal{F}_1$, $\mathcal{F}_2$ and $\mathcal{F}_4$ (See Fig~\ref{fig:finetuning_matrices}.). In addition, clusters 2 and 4 were visually different as seen in Fig~\ref{fig:cluster_analysis_pathmnist} with cluster 4 corresponding to darker less textured images.
%
Finally, only cluster 8 seemed to contain \textit{normal colon mucosa} (in addition to \textit{Debris}) and does seem to contain darker textured images than other false positive clusters.

\begin{figure}[ht!]
    \centering
    
    \includegraphics[width=\linewidth]{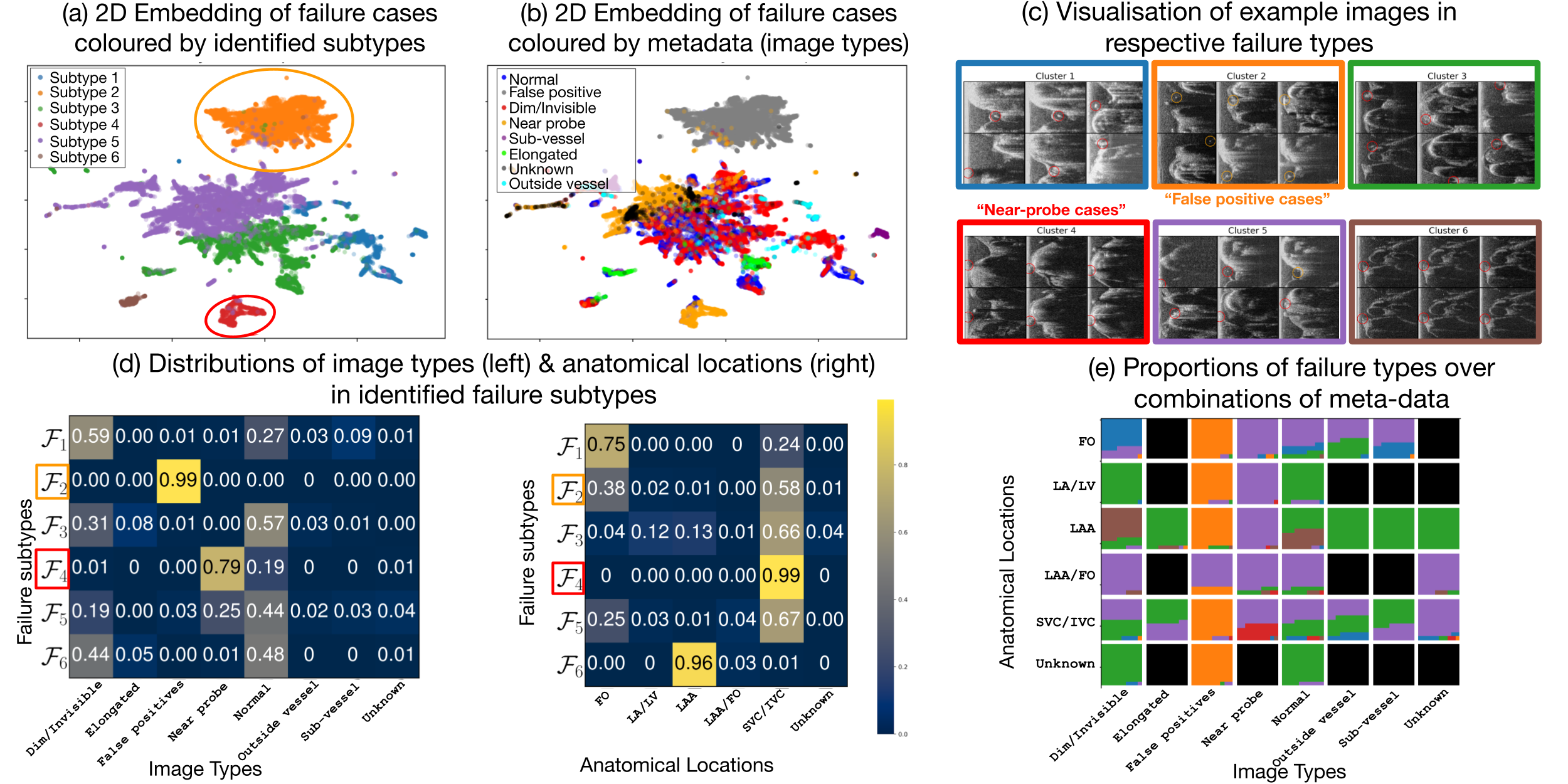}
    
    \caption{Inspection of the failure types obtained through gradient space clustering for ICE-CD. UMAP embedding coloured by failure types (a) and by image types (b). (c) Example images for each subtype.  (d) Distribution of image types and anatomical locations within each subtype.  (e) Proportions of failure types over each combination of anatomical location and image types coloured by failure type.
    Two outstanding failure types are $\mathcal{F}_2$ focusing on ``false positives'' and $\mathcal{F}_4$ focusing on ``Near-probe'' cases.
    }
    \label{fig:cluster_analysis_caths}

\end{figure}

\textbf{Catheter detection (ICE-CD):} \textit{Gradient clustering} was able to recover a known and important but under-represented failure type: Cluster 4 (red cluster in Fig.~\ref{fig:cluster_analysis_caths}(a)) focused on \textit{Near-probe} catheters which are close to the ultrasound probe. Indeed, these catheters are hard to detect due to noise in this region of the images.
Secondly, \textit{Gradient clustering} was able to automatically discover some of the anatomical locations.
Indeed, Cluster 6 (brown cluster in Fig.~\ref{fig:cluster_analysis_caths}(a)) focused on the \textit{LAA} and Cluster 3 (green cluster in Fig.~\ref{fig:cluster_analysis_caths}(a)) focused mostly on the \textit{SVC/IVC}.
Finally, \textit{Gradient clustering} was able to separate false positives (see the orange cluster in Fig.~\ref{fig:cluster_analysis_caths}(a)) from false negatives (the others).

\subsection{Model Repairment}

\textbf{Experiments:} we aim to evaluate how much failure types performance can be improved while retaining performance on correct cases (See Table~\ref{tab:model_repair}). 
We compare several repairment approaches on both datasets based on fine-tuning and EWC~\cite{kirkpatrick2017overcoming}. The fine-tuning is done on either a single failure type: $\mathcal{F}_i^{tr}$ or all: $\mathcal{F} = \cup_i \mathcal{F}_i^{tr}$. 
Also, we compare to using $\mathcal{C}^{tr}$ with a 50\% sample ratio.
For all methods, early stopping is performed by selecting the best accuracy on $\mathcal{F}_i^{tr}$. Models were fine-tuned with a learning rate of $10^{-6}$ on BP-MNIST and $10^{-4}$ on ICE-CD and weight decay of $10^{-3}$.  
Table~\ref{tab:model_repair} reports the accuracy on the test set of each failure type  $\mathcal{F}_i^{te}$, the previously correct cases $\mathcal{C}^{te}$ and the overall test set.

\textbf{Analysis:} first, fine-tuning on a single failure type $\mathcal{F}_i^{tr}$ generalized more on that specific failure type than fine-tuning on all incorrect cases $\mathcal{F}^{tr}$  at once (see Table~\ref{tab:model_repair}).
This may indicate that the failure types are conflicting during fine-tuning and it is more difficult to simultaneously learn a diverse set of cases than simple ones.
Therefore, if learning unimportant failures is conflicting with critical ones, it makes sense to first start by repairing a carefully selected sub-set of the failures.
Secondly, fine-tuning on the failures only couldn't maintain performance on $\mathcal{C}^{te}$ while including the correct cases $\mathcal{C}^{tr}$ helped to preserve performance. Fine-tuning on $\mathcal{F}_i^{tr}\cup \mathcal{C}^{tr}$ for ICE-CD dropped to 0.73 but this was still higher than 0.32 if using only $\mathcal{F}_i^{tr}$ for fine-tuning.
Finally, while for EWC~\cite{kirkpatrick2017overcoming} the performance on correct cases didn't drop as much as simple fine-tuning, EWC wasn't able to maintain correct cases accuracy to more than 0.27 and 0.51.

\begin{table}[ht!]
    \caption{Comparison of model repairment methods on BP-MNIST (top) and ICE-CD (bottom) evaluated on failure types obtained in gradient space (ACC). Methods retaining at least 0.85 accuracy on $\mathcal{C}$ are shown in green and those that do not in red.}
    \label{tab:model_repair}
        \centering
    \begin{tabular}{|c|c|c|c|c|c|c|c|c|c|c|}
    \hline
    
    Method (BP-MNIST)  & $\mathcal{F}_1^{te}$ & $\mathcal{F}_2^{te}$ & $\mathcal{F}_3^{te}$ & $\mathcal{F}_4^{te}$ & $\mathcal{F}_5^{te}$ & $\mathcal{F}_6^{te}$ & $\mathcal{F}_7^{te}$ & $\mathcal{F}_8^{te}$ & $\mathcal{C}^{te}$ & All \\
    \hline
    
    Pre-repairment  & 0 & 0 & 0 & 0 & 0 & 0 & 0 & 0 & \cellcolor{green!15}1.0 & 0.91 \\

    Fine-tuning on a single $\mathcal{F}_i^{tr}$   & 0.92 & 0.94 & 0.97 & 1.00 & 0.99 & 0.94 &  0.96 & 0.92 & \cellcolor{red!15}0.81 & 0.77 \\
    
     Fine-tuning on $\mathcal{F}_i^{tr} \cup \mathcal{C}^{tr}$  & 0.91 & 0.93 & 0.92 & 0.97 & 0.99 & 0.89 & 0.88 & 0.89 & \cellcolor{green!15}0.91 & 0.86 \\
     
    \hline
    
    Fine-tuning on $\mathcal{F} = \cup_i \mathcal{F}_i^{tr}$  & 0.4 & 0.79 & 0.72 & 0.83 & 0.87 & 0.71 & 0.82 & 0.78 & \cellcolor{red!15}0.19 & 0.24 \\

    EWC~\cite{kirkpatrick2017overcoming} on $\mathcal{F}^{tr}$  & 0.34 & 0.65 & 0.79 & 0.79 & 0.85 & 0.64 & 0.79 & 0.76 & \cellcolor{red!15}0.27 & 0.30 \\

    Fine-tuning on $\mathcal{F}^{tr}\cup\mathcal{C}^{tr}$  & 0.25 & 0.65 & 0.71 & 0.66 & 0.80 & 0.48 & 0.69 & 0.68 & \cellcolor{green!15}0.9 & 0.88 \\
    
    \hline
    \hline
    
    \hline
    
    Method (ICE-CD) & $\mathcal{F}_1^{te}$ & $\mathcal{F}_2^{te}$ & $\mathcal{F}_3^{te}$ & $\mathcal{F}_4^{te}$ & $\mathcal{F}_5^{te}$ & $\mathcal{F}_6^{te}$ & - & - & $\mathcal{C}^{te}$ & All \\
    \hline
    
    Pre-repairment & 0 & 0 & 0 & 0 & 0 & 0 & - & -  & \cellcolor{green!15}1 & 0.59 \\

    Fine-tuning on a single $\mathcal{F}_i^{tr}$   &  0.85 &  0.99 & 0.77 & 0.69 &  0.64 &  0.96 & - & - & \cellcolor{red!15}0.32 & 0.36 \\
    
     Fine-tuning on $\mathcal{F}_i^{tr}\cup \mathcal{C}^{tr}$ & 0.81  &  0.97 & 0.79 & 0.96 & 0.67 & 0.97 & - & - & \cellcolor{red!15}0.73 & 0.64  \\
    
    \hline
    
    Fine-tuning on $\mathcal{F} = \cup_i \mathcal{F}_i^{tr}$   & 0.61 & 0.9 & 0.71 & 0.28 & 0.57 & 0.72 & - & - & \cellcolor{red!15}0.64 & 0.66 \\

    EWC~\cite{kirkpatrick2017overcoming} on $\mathcal{F}^{tr}$  & 0.41 & 0.89 & 0.59 & 0.26 & 0.47 & 0.94 & - & - & \cellcolor{red!15}0.51 & 0.56 \\
    
    Fine-tuning $\mathcal{F}^{tr} \cup \mathcal{C}^{tr}$  & 0.41 & 0.8 & 0.65 & 0.39 & 0.61 & 0.82 & - & - & \cellcolor{green!15}0.86 & 0.78 \\
    
    \hline
    
    \end{tabular}

\end{table}


\section{Conclusion}

We have introduced a principled framework to address the problems of failure identification, analysis and model repairment.
Firstly, we put forward a set of desirable properties for meaningful failure types and novel surrogate metrics to assess those properties in practice.
Secondly, we argued that model repairment should not only aim to fix the failures but also to retain performance on the previously correct data.
Finally, we showed specific instantiations of our framework and demonstrated that clustering in feature and gradient space can automatically identify clinically important failures and outperform manual inspection.

%
%
%
\bibliographystyle{unsrt}
\bibliography{paper1329}

\end{document}